\documentclass[conference]{IEEEtran}
\IEEEoverridecommandlockouts
\usepackage{cite}
\usepackage{amsmath,amssymb,amsfonts}
\usepackage{algorithmic}
\usepackage{graphicx}
\usepackage{textcomp}
\usepackage{xcolor}
\def\BibTeX{{\rm B\kern-.05em{\sc i\kern-.025em b}\kern-.08em
    T\kern-.1667em\lower.7ex\hbox{E}\kern-.125emX}}

\newcommand{\tensor}[0]{\otimes}

\newcommand{\Tensor}[2]{#1{\tensor}#2}

\newcommand{\R}{\mathbb{R}}

\newcommand{\Vc}[1]{{\boldsymbol #1}}

\newcommand{\singlefigure}[4]{{\begin{figure}[htb] %
\centering %
\includegraphics[width=#1\linewidth]{#2}%
\caption{#3}%
\label{#4}%
\end{figure}}}

\begin{document}

\title{DPUV3INT8: A Compiler View to programmable FPGA Inference
  Engines}

\author{
  \IEEEauthorblockN{ Paolo D'Alberto}
  \and
  \IEEEauthorblockN{ Jiangsha Ma}  
  \and
  \IEEEauthorblockN{ Jintao Li}  
  \and
  \IEEEauthorblockN{ Yiming Hu}  
  \and
  \IEEEauthorblockN{ Manasa Bollavaram}  
  \and
  \IEEEauthorblockN{ Shaoxia Fang}
}

\maketitle

\begin{abstract}
We have a FPGA design, we make it fast, efficient, and tested for a
few important examples. Now we must infer a general solution to deploy
in the data center. Here, we describe the FPGA DPUV3INT8 design and
our compiler effort.  The hand-tuned SW-HW solution for Resnet50\_v1
has (close to) 2 times better images per second (throughput) than our
best FPGA implementation; the compiler generalizes the hand written
techniques achieving about 1.5 times better performance for the same
example, the compiler generalizes the optimizations to a model zoo of
networks, and it achieves 80+\% HW efficiency.
\end{abstract}


\section{Introduction}
\label{sec:introduction}

We shall present technical and common-practice arguments in the
compiler field: for example, loop tiling, software pipelining, and
code generations. In the literature, there are very few examples about
compilers and their role in facilitating acceleration by custom
hardware. However, accelerators are ubiquitous and custom hardware are
becoming common in AI. TVM is an example where compiler approaches are
applied towards such ends \cite{ChenMSYWHCGK2018}. Also the research
in the field present a plethora of optimizations at different level of
the code generation stack: from formula manipulation (like FFT
\cite{PuschelFV11,FrigoJ98}), to thread allocation (MAGMA and Cilk
\cite{GaneshanEA20,FrigoHLL09}) to scheduling
\cite{KejariwalVNGTS09}. The application of accelerators is not new
and OpenCL \cite{PaulinoFC20,Nugteren17}, SYCL \cite{LalASCHSTFH20}
provided heterogenous solutions and more proprietary solution such as
CUDA \cite{AlMouhamedKM20} and ROCM \cite{rocm} are dominating GPUs
use, where the compiler magic is done per GPU. More often than not
they are based on general solutions \cite{LattnerA2004} and their
applications. Often, the application has to be {\em rewritten} not
just re-compiled (like we could \cite{dalberto2015multiplecampaign}),
or at least we need to tailor kernels, so that a general-purpose
compiler will not do. So where are we?

Our HW is simple, fast, and tailor made by combining low latency and
higher throughput. The HW simplicity requires the compiler working
harder. FPGA based architectures are designed to do a few instructions
extremely well (i.e., correlation) and no attention is given to any
other (i.e., softmax).  Foremost, this creates partitioning hurdles to
their deployments. Also, in this work such operations are basically
custom instructions and they are not {\em closed}: for example, one
convolution HW instruction cannot compute and cannot describe a wholesome
convolution layer.  The number and complexity of these instructions
are a function of the I/O sizes, weight size, and often the
context. In practice, compilers are harder to write and inevitable
development tools.

Code optimizations and generation is context dependent: A few
important considerations and examples follow.
\begin{itemize}
\item For example tiling: here, not every tile size is valid because
  of HW constraints, a cost function is not often available, the size
  of the tile is based on memory foot print more than data dependency
  and operation costs, the size of the loop polyhedron can be very
  large, we have hundreds of them, optimal solutions are easily
  written by hand only if the network is simple and the size is fixed,
  the steady computation is often different from the prologue and
  epilogue thanks to padding and strides or by fusing layers together
  into a single operation.

  For emphasis, a convolution is a single layer, its execution is a
  composition of elementary operations (HW instructions), and they
  depend on the context. Convolution and pooling can be merged into a
  super layer; the partitioning of the super computation may not be
  better than the two separately.  In practice, data dependency tools
  for the computation using distance vectors can be a little hard to
  apply and the loops themselves may not look so well defined.

\item We are in a scenario where parallelism is more like thread
  parallelism, memory allocation is per thread but it is a function of
  the computation schedule and there is no hardware support for any
  thing than a few computational instructions. The least understood
  problem is that tensors (variables or activations) do not have
  neither standard layout and they have different layouts in different
  memory levels. The ability to represent tensors as a single memory
  space is a property we appreciate fully only when it is missing.

\item We deploy two types of loop tiling, two-layer fusion, and
  software pipelining in a scenario where the loop sizes would not be
  practical (because of inherent complexity) for such advanced
  compiler optimizations. Here these optimizations are necessary to
  exploit the HW potential.
\end{itemize}

In the following, we show how a compiler code generation exploits near
peak efficiency 80+\% for a new generation of FPGA based custom
architectures. 

\section{Directed Acyclic Graph: computational model}
\label{sec:dag}
\singlefigure{0.95}{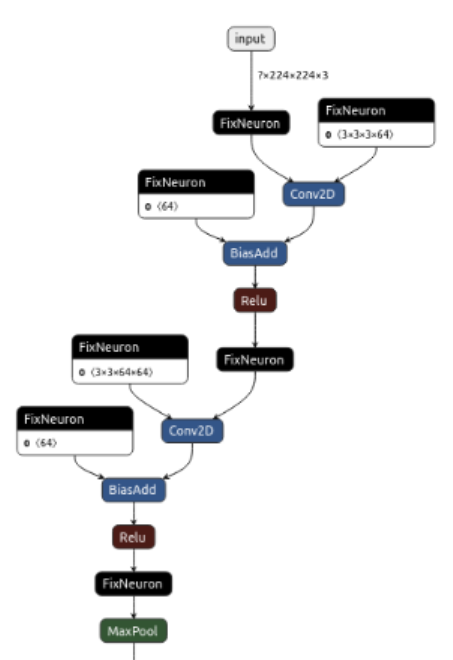}{VGG 16 first three layers:
  Conv1+Relu + Conv2+Relu + MaxPool }{fig:network}

In general, we have a tensor as input, say an image. We have multiple
outputs: an image and annotations.  We represent the computation by a
direct acyclic graph (DAG). A node represents a computation with
multiple inputs and a single output. We separate inputs and
parameters. Parameters are constants and not computed: they are read
(once) and from a dedicated memory space. Each node has a single
output tensor and each node is connected by a path to both an input
and an output node. Tensors are up to four dimensional vectors and
they represent data and a {\em memory} space. A node is associated
with a single output tensor (e.g., written once and read by many
nodes) and at least one input tensor (e.g., written by a single node,
unless the node is an input itself). The DAG and a topological ordered
sequence of nodes, a {\em schedule}, represent a valid computation and
a computation is a sequence of Static Single Assignment statements
(SSA). A graph may have multiple inputs and multiple outputs, a graph
node has at least one input and only one output.

In Figure \ref{fig:network}, we present the {\em netron} graphical
representation \cite{netron} of a enhanced and quantized tensorflow
graph. We represent two convolutions and one {\em maxpool}. There are
custom {\em FixNeuron} layers as quantization transformation layers
for weights, biases, and for activation tensors so that to transform
the original float arithmetic into 8 bits. Weight and bias are
parameters and read only once. FixNeurons contain the information
about range and precision of the result tensors and it is a
quantization solution used within the Vitis framework
\cite{vitis-ai}. Such a graph structure is integral part of the front
end (i.e., quantization and it may involves re-training or fine
tuning) it will assure that the computation can be carried on using
8bit arithmetic.  Parameters and neurons are assimilated by the
compiler into the corresponding node (i.e., convolution) in order to
reduce the size of the graph and its complexity. Eventually, we reduce
the 15 nodes to 4: 1 input, 2 convolutions, and 1 maxpool (and
optimized to 3 by fusing convolution and maxpool). The schedule is
shorter. These graph optimizations are standard nowadays and we do not
dwell on them. Also, we assume that the DAG has operations we can
reduce to basic FPGA operations (e.g., no softmax no tensor reshapes).

We present a recursive approach that divides/tiles the node operations
in custom and optimized elementary instructions. The memory hierarchy
and the schedule specify when and where tensors are computed and
stored; this summarizes the memory foot print and memory pressure; in
turn, the available space determines potential parallelism and
feasible scheduling. The compiler has tools for the exploration of
schedule and parallelism exploitation by splitting the graph into
small enough convex sub-graphs (i.e., the sub-graph has no path
exiting and re-entering the subgraph). We shall not present details
about these tools at this time (you may find more details about memory
allocation and schedule exploration in \cite{DAlbertoWNNDS2021}). The
elementary instructions are specific to the hardware thus we must
present at least its abstraction.

\section{Hardware Abstraction}
\label{sec:hwabstraction}

\singlefigure{0.95}{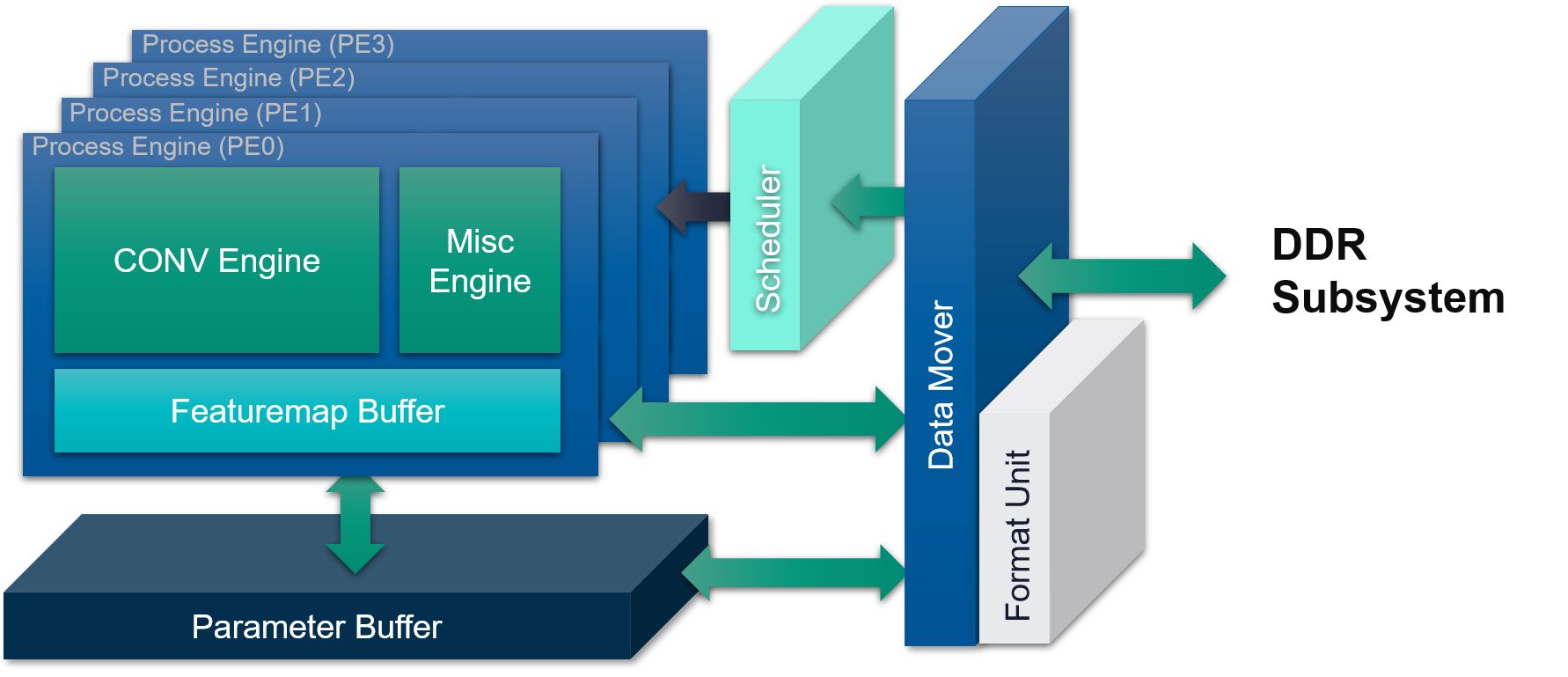}{Basic HW abstraction}{fig:hardware}

We present a graphical abstraction of the hardware in Figure
\ref{fig:hardware}.  We have a main DDR memory where Inputs, Outputs,
Parameters, and Instructions are stored. We have one main unit
composed by four Process Engine (PE) and one Parameter memory (PM) for
parameters. Every PE has a Feature map memory (FM) for input and
output tensors and two computational engines: CONV and MISC. The PEs
represent a Single Instruction Multiple Data (SIMD) architecture and
it will work computing 4 tensors at any time (up to 16 concurrently on
a U280). For compiler purposes, the code generation focus on a single
PE and thus a single tensor computation, and only for data movement
to/from DDR multiple tensors are involved (i.e., 4 tensors).

The memories PM, FM, and DDR are scratch pads. The compiler has the
responsibility to manage allocation and de-allocation of every
tensors. The PM and FM are circular buffers: that is a tensor can wrap
around the last address back to the beginning and a tensor is stored
contiguously. The circular property is necessary for streaming data to
and from DDR where only a part of the tensor will be transient but
contiguous and without overlapping. The DDR is not circular, we assume
tensors will fit entirely, and it is logically split by register
pointers into 5 parts: inputs, outputs, parameters, instructions, and
swap/stack for memory data management (the equivalent of a compiler
stack or {\em malloc} space). The architecture is more efficient for
tensors with no small dimensions we call {\em channels}. The Format
Unit is designed to change the layout of the inputs (images) on the
fly to improve efficiency, due to the small number of 3 RGB channels
(i.e., images and image classification). If the number of channels is
larger than 8, the format unit is circumvented.

Every PE has a FM and four functional units (two shared).  The FM is
composed of three distinct circular memories with one read and one
write port each; every memory is composed of eight banks that we can
read and write them in parallel.  The four functional units are:

\begin{itemize}
\item LOAD: This moves data from DDR to FM by means of a LOAD
  instruction. The same instruction can move parameters from DDR to PM.
\item SAVE: This moves data from FM to DDR.
\item CONV: This is a systolic array convolution engine: it reads the
  parameters from the common PM, the inputs and the outputs from its
  local FM.
\item MISC: The miscellaneous unit is responsible for three types of
  operations: max pool, element wise addition, and data movement used
  for sample, up-sample, and identity. It reads and writes in its local
  FM.
\end{itemize}

\singlefigure{0.95}{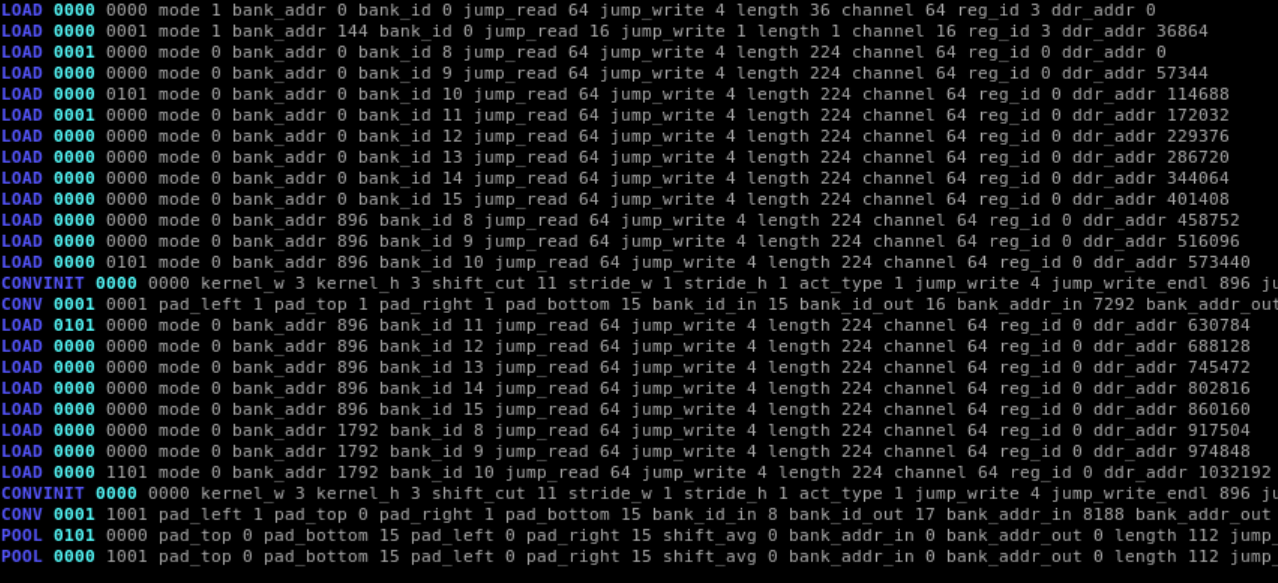}{Sample of our human-readable code}{fig:code}
There is a correspondent instruction for each unit: LOAD, SAVE, CONV,
MISC (MAXPOOL and ELT). Each unit can execute in parallel and thus
each instruction can be issued in parallel. Each instruction is
synchronized by two entries: DPON and DBBY. DPON describes the
operation types the current instruction depends on (multiple
instructions). DPBY describes the operation types the current
instruction depend by (multiple instructions).  The dependency is by
type only. An excerpt of generated code (without comments) is in
Figure \ref{fig:code}, and the first two entry of the instruction
represent the instruction dependency DBON and DPBY (i.e.,
\verb2CONV DBON DBBY2 ...). The specific value of the dependency
encoding is immaterial at this time. However, if you are thinking
about issuing instruction sequentially these attributes will be
associated with the preceding and succeeding instruction in the
schedule. This is easy.  If you are thinking about parallelism, how to
issues the instructions so that these dependencies will rely only on
the type of the instructions is not that easy. The compiler cannot
rely on when nor on the instruction number since issued instruction
0. For example, software pipelining is not straightforward. We shall
return and explain this in Section \ref{sec:pipelining}.

\section{Nodes and Computations}
\label{sec:node}
Consider a convolution operation $\Vc{Y} =
\Tensor{\Vc{W}}{\Vc{X}}+\Vc{B}$, where $\Vc{X} \in \R^{h_i,w_i,c_i}$,
$\Vc{Y} \in \R^{h_o,w_o,c_o}$, and $\Vc{W} \in
\R^{c_o,h_L,w_M,c_i}$. The full definition is
\begin{multline}
  \label{eq:3}
  \Vc{Y} \leftarrow {\dot\sum}_{i=1}^{h_o}{\dot\sum}_{j=1}^{w_o} {\dot\sum}_{k=1}^{c_o} \\
  \sum_{l=1}^{h_L}\sum_{m=1}^{w_L} \sum_{n=1}^{c_i} x_{(i+l-1,j+m-1, n)}  w_{(k,l,m,n)}+b_k\large
\end{multline}

Any activation tensor has three main dimensions: the channel $c_i$,
which is the innermost; the width $w_i$, which is the middle one; and
the height $h_i$, which is the outermost. The tensor is stored in
memory following the same logic. Here, we omit any batch size
reference.

Each element of the $\Vc{Y}$ tensor is the following computation based
on summations:
\begin{equation}
  \label{eq:1}
  y_{(i,j,k)} = \sum_{l=1}^{L}\sum_{m=1}^{M} \sum_{n=1}^{N} x_{(i+l-1,j+m-1, n)}
  w_{(k,l,m,n)}+b_k
\end{equation}
We refer to this computation as a {\bf convolution}. A convolution is
defined as a 6 dimensional loop, Equation \ref{eq:1}, represents the
innermost loop and the outer loop is three iteration loop
\begin{equation}
  \label{eq:2}
  \Vc{Y} \leftarrow \dot\sum_{i}^{h_o}\dot\sum_{j}^{w_o}
  \dot\sum_{k}^{c_o} y_{(i,j,k)}
\end{equation}

From here on, we will not specify $\dot\sum$, that is an iterative
loop, and use the $\sum$ symbol everywhere; any misunderstanding will
not be harmful and the formulation concise.  The computation will go
from right to left and we write the formulae from left to right.

The process of thinking of $\Tensor{\Vc{W}}{\Vc{X}}+\Vc{B}$ as a
single operation like a loop, partitioning it into smaller loops, and
then reconstruct as a unfolded sequence of basic convolutions is a
useful one. We use a declarative expression using loops to show the
details of the computation and express the division of the
computation. Programmatically, the recursive nature is simpler to
use. In our heads, when we talk about tiling we mean recursion. In our
framework, there is no Loop construct, and a computation is a tree
where each leaf is a single computation summarized by a few HW
instructions and inner nodes represent sub-computations (and data
movements). However, recursion makes a poor presentation tool to the
compiler community at large where iterative loops are dominant.
Clearly, having 9 or more loops as code description will give too much
information and will be cumbersome. We decided for presentation sake
that will play with a mixed representation to keep the iterative
nature of the computation and the recursive division and code
generation (to keep it short). Hopefully, the mixed approach will
strike an advantageous compromise (or make everybody unhappy).

\section{The Recursive Description of a Computation (into Computations)}
\label{sec:computation}

A techincal distinction is in order. A node in the DAG is a fully
defined computation and it is atomic. The main point to realize is the
computation of a node is a function of the context in a DAG (this is
obvious). For a node computation, the input addresses and the output
address are specified, the properties of the node like stride and
padding are associated to the computation. However, if we can divide
the computation into components, each component is a computation that
may have different attributes (i.e., padding) and access input and
outputs with patterns that are not the same of the original. For
example, let us consider Equation \ref{eq:1}, the order of the loops
determines the input access patterns, the sub computation may have to
exploit different ones from the original.

The recursive description of code generation follow this idea from the
largest to the smallest problem: We split the computation by width, by
height, and last by weights. We present the process using convolutions
but the same idea apply for all computations (well not the weight
decomposition)

\subsection{W-splitting: overlapping inputs}
Consider again the convolution computation:
$\Vc{Y}=\Tensor{\Vc{W}}{\Vc{X}}+\Vc{B}$ with unitary stride, no
padding. We split the tensor $\Vc{Y}$ in two parts by the width
dimension $\Vc{Y}_{i=1 \equiv [1,a]}$ and $\Vc{Y}_{i=2 \equiv
  [a+1,w_o]}$:
\begin{equation}
  \label{eq:wsplitting}
  \Vc{Y} = \Vc{Y}_{1} | \Vc{Y}_{2} = \sum_{v=1}^2\Vc{Y}_v
\end{equation}
To perform the computation we need to split the input vector into two
parts partially overlapping:
\begin{equation}
  \Vc{X} =  \Vc{X}_{w_i \in [1,(a-1)+L]} | \Vc{X}_{w_i \in [a,w_i]} 
\end{equation}
There is an overlap if $L>1$, see Figure \ref{fig:overlap}
\singlefigure{0.5}{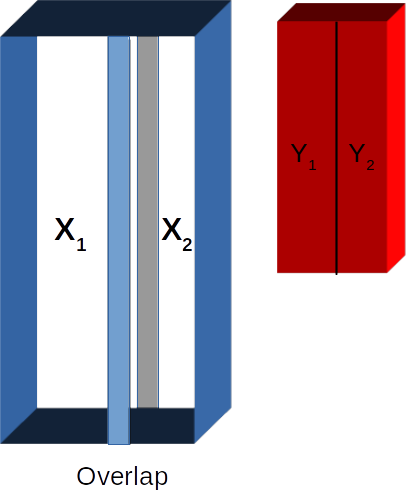}{W-Splitting with input overlap azure
  band between $X_1$ and $X_2$}{fig:overlap}
\begin{equation}
  \Vc{Y} =  \sum_{v=1}^2 \Tensor{\Vc{W}}{\Vc{X}_v}+\Vc{B}
\end{equation}
If there is a stride $s_w$ and padding $p_w$:
\begin{equation}
  \Vc{X} =  \Vc{X}_{w_i \in [1,(a-1)s_w+L]} | \Vc{X}_{w_i \in [a*s_w,w_i]} \\
\end{equation}
\begin{equation}
  \Vc{X} =  \Vc{X}_{w_i \in [1,(a-1)s_w+L-p_w]} | \Vc{X}_{w_i \in [a*s_w-p_w,w_i]} 
\end{equation}
The complete formulation is cluttered, we shall omit it for simplicity
whenever possible. Consider a tensor of size $1\times w \times c$, our
hardware has a few requirements about the size of this simple size
tensor. If we say that $\gamma$ is $\max(w*c)$ for which we can read
and perform operation on the tensor, we split the computation so that
\begin{equation}
  \Vc{Y} =  \sum_{v=1}^{\lceil\frac{w*c}{ \gamma} \rceil} \Tensor{\Vc{W}}{\Vc{X}_v}+\Vc{B}
\end{equation}
The data dependency between the input and the output (by the weight
tensor shape) forces an overlapping space between iterations. As a
matter of fact, the computations are independent and Equation
\ref{eq:wsplitting} shows this is the first recursive level and it is
possible only if either input or output have $\max(w*c)\geq
\gamma$. Longer vector in the width dimension will take more
space. Thus less space to perform double buffering using the maximum
parallelism of the architecture. The W-Splitting is the classic loop
tiling only that tiles share inputs points, tensors are meant to be in
DDR, $X_1$ and $X_2$ are scattered in DDR, but otherwise vectors will
stream well in FM memory where they will be contiguous: The
H-Splitting is tiling and software scheduling so that to optimize data
streaming and architecture parallelism.

\subsection{H-Splitting}
As usual, we consider $\Tensor{\Vc{W}}{\Vc{X}}+\Vc{b}$ with $\Vc{Y}
\in \R^{h_o,w_o, c_o}$. We have a preferred computation height $H_C
=8$ for convolution ($H_E, H_P =2$ for addition and pool). This means:
first, we have a preference in producing an output tensor with height
$H_c$; second, smaller dimension are possible but less efficient;
third, the tiling is function of the operation type (and if they are
fused, not straightforward).  If $\Vc{Y}_v$ means $y_{i,*,*}$ with $i
\in[(v-1)H_c+1, vH_c]$ then we can use the previous notation to
describe the computation:

\begin{equation}
  \Vc{Y} =  \sum_{v=1}^{\lceil\frac{h_o}{ H_c} \rceil} \Vc{Y}_v = \sum_{v=1}^{\lceil\frac{h_o}{ H_c} \rceil} \Tensor{\Vc{W}}{\Vc{X}_{v}} +\Vc{B}
\end{equation}

The minimum output foot print is $(H_c,W_o,C_o)$, the input has foot
print with stride $s_h$ and hernel $k_h$
\begin{equation}
  Size(\Vc{X}_v) = ((H_c-1)*s_h + k_h , W_i, C_i)
\end{equation}
If the minimum foot print does not fit any of the memories in FM, we
shall reduce $H$ accordingly. Of course, smaller $H$ will produce less
efficient computations and slower executions. At this stage, we
reformulate the computation so that we have $\frac{h_o}{H_c}$
convolutions and each will compute a $H_c$-tall output tensor. If
reducing the height to one is not enough, we must w-split the
computation first.

As you can see, W-Splitting and H-Splitting are very similar. There
are a few differences:
\begin{enumerate}
 \item The W-splitting is triggered by the number of available rows in
   the feature-buffer banks (FM) 
 \item The H-splitting is triggered by the number of banks in FM and
   the computational parallelism.
 \item H-Splitting will not have input overlap (we shall explain
   further) by exploiting streaming and computational parallelism.
\end{enumerate}

\section{Operation Fusion}
\label{sec:fusion}
Our HW can perform in parallel pool and convolution operations. For
inference network it is common to fuse convolution and pool operations
into a single operation. In previous architectures, the main idea is
to reduce the memory pressure by removing the need of the intermediary
tensor (by using a smaller and temporary one). In this architecture we
are actually do fusion to exploit instruction parallelism and data
streaming. 
\begin{equation}
  \Vc{Y} = {\bf MaxPool} (\Vc{T} = \Tensor{\Vc{W}}{\Vc{X}} + \Vc{b})
\end{equation}
where using the summation as a loop:
\begin{equation}
  Y_{i,j,k} =  \sum_{n=1}^{k_w}\sum_{m=1}^{k_h}max((T_{i+m-1,j+n,k}) 
\end{equation}
We can tile by the usual height but the maximum height is $H_p=2$   
\begin{equation}
  \Vc{Y} = \sum_{v=1}^{\lceil\frac{h_o}{ H_p}\rceil} {\bf MaxPool}(\Vc{T}_v)
\end{equation}
The output sub-tensor will have foot print $(H_p, w_o, c_o)$, the
intermediary tensor, which is the output of the convolution, has a
foot print:  
\begin{equation}
  (T_h = (H_p-1)*PoolStride_h + PoolKernel_h , W_t, C_t)
\end{equation}
and the input
\begin{multline}
  ( (T_h-1)*ConvStride_h + ConvKernel_h , W_i, C_i) = \\
  (((H_p-1)*PS_h + PK_h-1)*CS_h +CK_h , W_i, C_i)
\end{multline}
\begin{equation}
  \label{eq:two-stages}
  \Vc{Y} = \sum_{v=1}^{\lceil\frac{h_o}{ H_p}\rceil} {\bf MaxPool}(
  \sum_{w=1}^{\lceil\frac{h_t}{ H_c}\rceil}(\Tensor{\Vc{W}}{\Vc{X}_w} + \Vc{b})_v)
\end{equation}
\begin{equation}
  \label{eq:fusion}
  \Vc{Y} = \sum_{w=1}^{\lceil\frac{h_t}{ H_c}\rceil}\sum_{v=1}^{\lceil\frac{H_c}{ H_p}\rceil} {\bf MaxPool}(
  (\Tensor{\Vc{W}}{\Vc{X}_w} + \Vc{b}))
\end{equation}
Equation \ref{eq:fusion} describes how the fusion operation works at
steady state. The formulation do not consider prologue nor epilogue,
there is no description of the effects of padding for input or
temporary space.  Clearly, it shows that instead of two passes as in
Equation \ref{eq:two-stages}, we have one pass of the input and of the
output and two computation stages.

Operation fusion has the side effect to create
tensors of different sizes and there is a multiplicative factor as a
function of the stride computation.  If the tile has to be reshaped to
accommodate smaller footprints, less efficient computations are
required and we turn off fusion. The same idea applies with little or
no modification to
\begin{equation}
  \label{eq:convpool}
  \Vc{Y} =  \Tensor{\Vc{W}}{{\bf MaxPool} (\Vc{X})} + \Vc{b}
\end{equation}


We call the steady state of the computation the number of convolution
output $H_c$ and the max pool output $H_p$. In practice, this is an
application of discrete mathematics: we have a valid steady state if
we can consume the convolution points at the same rate we can produce
them. That is, it exists a $k$ so that $kH_p= H_c$ and the
intermediate computation fits in memory.

In summary: W-Splitting, H-Splitting and fusion 
\begin{equation}
  \begin{split}
    \Vc{Y}  =  &   {\bf MaxPool}((\Tensor{\Vc{W}}{\Vc{X}} + \Vc{b}))     \\
    &   \sum_{v=1}^{\lceil\frac{w_i}{ \gamma} \rceil} {\bf MaxPool}((\Tensor{\Vc{W}}{\Vc{X}_v} + \Vc{b})) \\
    &  \text{ a.k.a.  W-Split}\\
    &   \sum_{v=1}^{\lceil\frac{w_i}{ \gamma} \rceil}\sum_{k=(v-1)*\gamma+1}^{v*\gamma} {\bf MaxPool}(\Vc{T}_k=(\Tensor{\Vc{W}}{\Vc{X}_v} + \Vc{b})) \\
    & \text{ a.k.a.  W-Split + fusion } \\
    &   \sum_{v=1}^{\lceil\frac{w_i}{ \gamma} \rceil} \sum_{k}\Big( \sum_{h=1}^{\lceil\frac{h_t}{ H_c}\rceil}\sum_{v=1}^{\lceil\frac{H_c}{ H_p}\rceil}  {\bf MaxPool}((\Tensor{\Vc{W}}{\Vc{V}_h} + \Vc{b})) \Big) \\
    & \text{ a.k.a. H-Split + fusion}\\
  \end{split}
\end{equation}
The notation becomes heavy, incorrect, difficult to read and to
understand. We will not indulge further notation describing W-Split
and H-split. The power of operation fusion is really evident when we
combine H-Splitting, operation fusion, and software pipelining.

\section{Software Pipelining as Code Reorganization}
\label{sec:pipelining}

Consider a node computation $\Vc{Y} = {\bf MaxPool} (\Vc{T} =
\Tensor{\Vc{W}}{\Vc{X}} + \Vc{b})$ where the input data is in DDR and
the output data will be in DDR, also consider the case where the PM
can fit the weights. The operation is fused and represented by
\begin{equation}
  \Vc{Y} = \sum_{w=1}^{\lceil\frac{h_t}{ 8}\rceil}\sum_{v=1}^{\lceil\frac{8}{2}\rceil} {\bf MaxPool}(
  (\Tensor{\Vc{W}}{\Vc{X}_w} + \Vc{b}))
\end{equation}
As a short and abstract description, the code for the computation
above: 
\begin{equation}
  \label{eq:abstractcode}
    {\bf Op}(\Vc{X}) =   \sum_{i=1}^{\lceil\frac{h_t}{ 8}\rceil} T_i \text{ and }
    T_i =   L_i  C_i  P_i  S_i    
\end{equation}
As a practical example for the first full computation of Convolution
and Pool: 
\begin{equation}
  \begin{split}
    T_1 = &  L_1  C_1  P_1  S_1     \\
    L_1 = &  \sum_{\ell=1}^{12} LOAD_{\ell} \text{, read 12-high vector}           \\
    C_1 = &  \sum_{c=1} CONV_{c}          \text{, 1 conv, 12 inputs,  8 outputs }\\
    P_1 = &  \sum_{p=1}^4 POOL_{p}        \text{, 4 pool,  8 inputs, 4 outputs}    \\
    S_1 = &  \sum_{s=1}^4 SAVE_{s}        \text{, 4 saves}
  \end{split}
\end{equation}

In general for $T_i$ and clearly visible in the particular $T_1$
instruction, each sub-computation has a different length of
instructions: 12, 1, 4, and 4 respectively. Also the execution time of
each sub computation is also function of the efficiency, shapes, and
size of the operations, and thus a function of how we split the
computation in parts. This is not a classic Very Long Instruction Word
(VLIW) although we want to exploit instruction level parallelism (it
is not VLIW because our different execution time of each
sub-computation implies that issuing the long instruction is not
enough to support their correct queuing and execution).

The tiling will split the computation into a sequence of basic
computations: because each tile has input data in DDR, the first
operation of the tile will be a load, then the main computation such
as convolution and pool (if it is a fused operation), and it will
conclude with a save instruction in order to move the data back into
DDR. In a practical context, each $L_i,C_i,P_i,$ and $S_i$ are a
sequence of instructions not necessarily a single instruction.

\singlefigure{0.7}{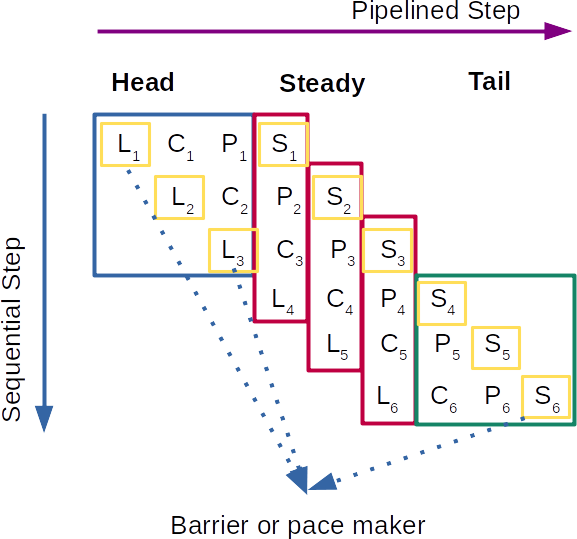}{Compiler software pipelining }{fig:sc}

As a reminder of the meaning of the our notation in Equation
\ref{eq:abstractcode}: we imply a strict dependency within the $T_i$
computation and we imply a dependecy from $T_i$ and $T_{i+1}$:
\begin{equation}
  L_1 \rightarrow C_1 \rightarrow P_1 \rightarrow S_1 \rightarrow   L_2 \rightarrow C_2 \rightarrow P_2 \rightarrow S_2 ...
\end{equation}
Thus the computation is purely sequential. The goal of software
pipelining is to unlock the parallelism from instruction $T_i$ and $T_{i+1}$.

The way we presented our computation in Section \ref{sec:computation},
we have the property that close-by tiles work on close-by data ($i$
and $i+1$) and they can stream into memory quite naturally. This has
the property that if we can exploit instruction parallelism we need
just to explore direct neighbors tiles so that to have the least
memory foot print. To exploit the parallelism we must reorganize the
computation differently and work on the dependencies.
\begin{equation}
  {\bf Op}(\Vc{X}) = Head + (Steady = \sum_{i=4}^{k} L_i, C_{i-1},
  P_{i-2}, S_{i-3}) + Tail
\end{equation}
where 
\begin{equation}
   Head = L_1 + (L_2 +C_1) + (L_3 + C_2 + P_1) 
\end{equation}
This means that $L_1, L_2, L_3$ are barrier instructions so that $C_1$
is parallel to $L_2$ and it can start only after $L_1$;  $L_3$ will
start only after $L_2$ and $C_1$ are done. Also notice that by the end
of the Head computation we have three live tensors $L_3$, $C_2$ and
$P_1$ that can be stored into a different FM.
\begin{equation}
  Steady = \sum_{i=4}^{k} (L_i, C_{i-1}, P_{i-2}, S_{i-3}) 
\end{equation}
$S_1$ will start only after $L_3, C_2 , P_1$, and $S_2$ after $L_4,
C_2 , P_2, S_1$. The $S_k$ are barrier or pace maker instructions. The
parentheses are used for emphasis to show that the instructions within
are parallel. The save operation $S_j$ is the only operation that will
free space (in the common sense). At steady pace, we can see the
parallel nature of the instruction issuing and that we are performing
a natural double buffering for each instruction. Take another look at
Figure \ref{fig:sc} and in particular at the instructions in {\em
  red}: $S_i$ instructions read contiguous sub-tensors and write them,
in practice only $S_i$ and $S_{i+1}$ need to be alive at the same time
to make sure the computation is coherent and correct. This is true for
the other operations.
\begin{equation}
   Tail =  (C_{k} + P_{k-1}+S_{k-2}) +(P_{k}+S_{k-1}) +(S_{k}) 
\end{equation}
Again $S_{k-2}, S_{k-1}, S_k$ are barriers and we can do that by the
proper use of DPBY and DPON:
\begin{equation}
  L_{i} + C_{i-1} + P_{i-2}, S_{i-3} :  DPBY  \rightarrow   S_{i-2}
\end{equation}
\begin{equation}
   L_{i} + C_{i-1}, P_{i-2}, S_{i-3} \leftarrow S_{i-2} :  DPON
\end{equation}

This is the basic idea of software pipelining for other architectures
and compilers: the main difference is we handle a sequence of
instructions instead of a single instruction common for long
vectorized instructions. Each instruction has different execution
time, the starting time is not assured by issuing the instruction but
the starting is function of the issuing of the previous instructions,
and synchronization with previous instructions is by type and thus by
DPBY and DPON events. Still the encoding of the dependency is
immaterial. If the memory space is a circular buffer, there is not
data overlapping.

{\bf Notes:} Software pipelining and circular buffers create a
streaming computation, a software driven streaming computation. The
barriers must change between the head computation and the rest because
by construction the $L_i$ will not be present in the tail and they may
be finished during the steady computation. There are corner cases
where we must introduce No-Operation bubble in the stream
computation. No-Op is a place holder for an instruction that is no
necessary anymore for the computation but it is necessary for the
smooth code generation.  The arguments here presented do not rely on
the order and length of the computation $T_i$; that is, $T_i =
L_i,P_i$ will work as above and it will exploit parallelism between
$L_i$ and $P_i$.

\section{Transpose convolution: optimal implementation}
\label{sec:deconvolution}

We know that the transpose convolution, known as deconvolution, is
equivalent to the sequence of an optional upsample and a
convolution. For our architecture, this is the only way to compute a
deconvolution. When there is an upsample, this means that the input of
the following convolution has lots of zeros. We avoid the zero
computation by reorganize the deconvolution into a set of
convolutions. In practice, we have a fully automated Deconvolution =
Convolutions transformation.

\singlefigure{0.7}{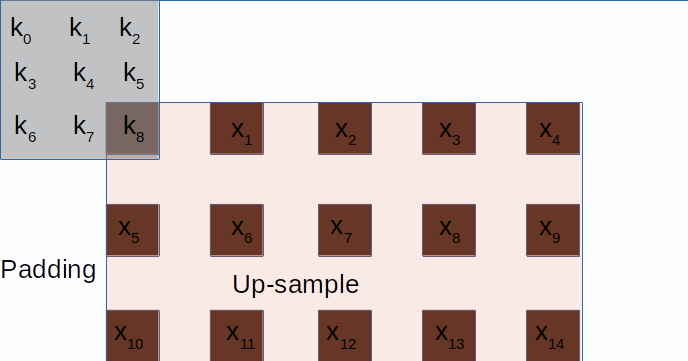}{Deconvolution= Upsample 2x2, Padding 2,
  kernel 3x3 }{fig:upandconv}

We describe the main idea by visualizing it for a deconvolution that
can be computed by a upsample 2x2, padding of 2, and a convolution
with kernel 3x3 and stride one. See Figure \ref{fig:upandconv}. If we
run the kernel, from the top left corner to the right, we see that the
first output is just the contribution input $x_0$ and $k_8$, The
second output is the contribution of $x_0$ and $k_7$, and the the
third output is $x_0*k_6 + x_1*k_8$. If we move the kernel one level
below and then to the right, we have $k_5*x_0$, $k_4*x_0$ and $k_3*x_0
+k_5*x_1$. Now let us take a look at Figure \ref{fig:series}

\singlefigure{0.9}{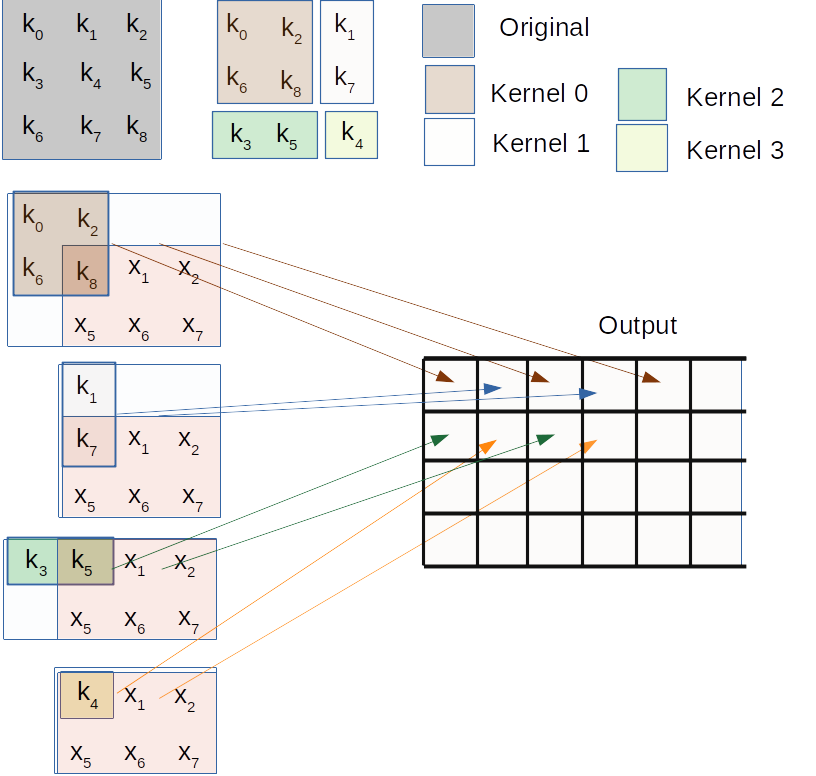}{Convolution Series =
  Deconvolution}{fig:series}

We can see that if we avoid to upsample the input and we reduce the
padding by 1, we can compute the same elements as before.  We take
$Kernel_0$, we set on the top left and the first output is $x_0*k_7$,
if we move the kernel by one spot to the right we have $k_6*x_0 +
k_8*x_1$ and this is the third output. So if we move $Kernel_0$ we
compute output $0,2,4,...$.  If we take $Kernel_1$ and we do not pad
left, we can compute output $1,3,5,..$. We take $Kernel_2$, we do not
pad the top of the input and we move the kernel to the right and we
compute the second row outputs $0,2,..$. Eventually we do not pad the
input and we use $Kernel_4$ to compute $1,3,5, ...$ of the second
row. We skip the details about the bottom and right padding for all
kernels because we need to do some book keeping and instead of clarity
we achieve likely the opposite.

The pictures above should be able to convince that the original
computation and the series of convolutions are equivalent. We need to
describe how we can systematically determine the kernels so that we
achieve such a property. The simplest way to visualize how the
sub-kernels are cut is by the overlapping of the original kernel and
the upsample input, like in Figure \ref{fig:kernels} but we do not
need to move the kernel around so much.
\singlefigure{0.5}{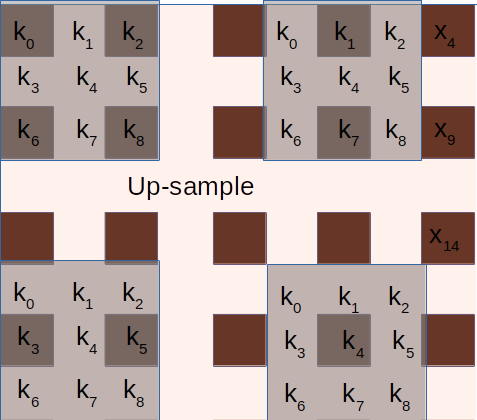}{Kernel determination = Original kernel and
  upsample pattern}{fig:kernels}

Given a kernel $k\times k$ and an upsample $m\times m$, there are
$m-1$ zeros, the number of kernels is $\min(k,s)^2$. One way to see
this intuitively, if $s=2$, the number of subkernels cannot be more
than 2 for any $k$. Here, we shall work with only upsample 2x2, and
thus the original kernel will be replaced by 4 subkernels. In this
section, we have the most variety as shape and size. For example, for
a kernel 4x4, we shall have 4 kernel 2x2.

The compiler allows to have a transpose convolution implemented as
$(2\times 2)$-Upsample + $k\times k$-Convolution or as 4 $(k-1)\times (k-1)$
convolutions.

\begin{equation}
  \label{eq:deconv}
  DECONV_{k,s,p}(X) = CONV_k(Padding_p(Upsample_s(X) ))
\end{equation} 
\begin{equation}
   \Tensor{\Vc{W}}{Pad(Usample(\Vc{X},s),p)} + \Vc{b}
\end{equation}
Pad is not an instruction, Upsample is a combination of MISC and LOAD
operations. The Upsample + convolution can be split and fused and we
can exploit software pipeline as in Equation \ref{eq:deconv1}
\begin{equation}
  \label{eq:deconv1}
   \sum_{v=1}^{\frac{H}{8}}
   \sum_{w=1}^{8}\Tensor{\Vc{W}}{Pad(Usample(\Vc{X}_w,s),p)} + \Vc{b}
\end{equation}
If we use the sequence of convolutions
\begin{equation}
  \label{eq:series}
   \sum_{v=1}^{\frac{H}{8*2}} \sum_{k=1}^{4}\Tensor{\Vc{W_k}}{Pad_k(\Vc{X}_v)} + \Vc{b}
\end{equation}
The equation is taking many liberties so that we can be concise. The
factor $8*2$ is because convolution has optimal output by $H_c = 8$,
but there are 2 {\em vertical kernels} such as $Kernel_0$ and
$Kernel_3$. We may have to reduce the $H_c$ to make sure we can
compute the output into feature map buffer. The size of $\Vc{X}_v$ is
similar to the one for regular convolution but not exactly: here we
require some book keeping and details will make the formulation
confusing.

As today, our architecture does not allow strided outputs by row, so
we need to perform the convolutions and then shuffle them by row
\begin{equation}
  \label{eq:series2}
   \sum_{v=1}^{\frac{H}{8*2}} \sum_{k=1}^{4}{Shuf}(\Tensor{\Vc{W_k}}{Pad_{k}(\Vc{X}_v}) + \Vc{b}))
\end{equation}
The shuffle is based on Element wise operations and thus can be done
in parallel with the convolution. So we keep the advantage of a
minimum number of computations and the formulation is symmetric to the
original Upsample + Convolution.

To show optimality: the series of convolution is a complete partition
of the original kernel without repetition. So every term in Equation
\ref{eq:series} is computed in Equation \ref{eq:deconv} as well,
however there is no zero computations and it cannot be done with any
fewer computations (without changing the definition or the
computation).

\section{Compiler Organization}
\label{sec:stack}

The compiler follows an organization similar to other compilers in the
field and within Xilinx's open software projects. 
\begin{enumerate}

\item We parse a network into an intermediate structure XIR composed
  by Op and Tensor classes (from Caffe, TensorFlow, and XModel --XIR)
\item We create a Hyper Graph (V,E) where a node in V is a (Op,Tensor)
  pair.  The Hyper Graph is designed so that the compiler can
  \begin{itemize}
  \item explore different schedules
  \item introduce operation fusion at graph level
  \item determine input and output nodes (thus interface)
  \item cut and reorganize the graph (i.e., debug)
  \item add nodes (i.e., memory coherence)
  \end{itemize}
\item We redirect parameters to be constants instead of operands
\item We enrich the tensors with quantization information:
  quantization information is associated with range, like $(-8, 8)$
  and minimum interval between two consecutive number in the range.
\item We perform graph manipulations in order to keep the minimum
  number of nodes.
\item Optional, we explore operation parallelism by scheduling and
  memory space estimation.
\item For every Schedule until success:
  \begin{itemize}
  \item we assign memory properties such as data memory layout
  \item we determine when and if operation fusion can be done
    efficiently
  \item we determine the memory allocation and reuse of inputs/outputs
    and parameters
  \item we identify the nodes where we can perform parameters
    prefetching into PM.
  \item we start the code generation of the primary computation by our
    recursive code generation
  \item streaming and software pipelining
  \end{itemize}
  
\end{enumerate}

Given a graph, the compiler creates two outputs: the instruction set
and the layout of the parameters in memory. We estimate and validate
the instruction dependency. The compiler can be a standalone
entity. In practice, it is used in combination with a partitioner: the
main graph is split and any subgraph related to our architecture is
the input to the compiler. The partitioner takes the output and
annotates the subgraph with code and parameters in binary format. The
run-time executes the partitioned graph (we will expand when showing
performance of the model zoo).

\section{Experimental Results}
\label{sec:experimental}
We present a graphical representation of the execution by an HW
simulator and the FPGA execution runs confirm the total execution time
accuracy (i.e. Figure \ref{fig:resnet}). Given the assembly code, it
presents the execution time of each instruction and it shows the
schedule and its execution. The architecture has four queues:
\begin{itemize}
  \item LOAD: weight and activation tensors move from DDR to FM (blue)
    and to PM (black). The Figure represents the execution time for
    the transfer by color accordingly.
  \item MISC: the operation maxpool (green) and Element wise one or
    two operands (yellow).
  \item CONV: operations like initialization (red) and convolution
    (purple) \
  \item SAVE: data movement from FM to DDR (purple).
\end{itemize}
The x-coordinate has double duties: it represents when an hw
instruction is issued and the thickness of the line tells the
duration. The x-coordinate represents time. The tools actually allows
to use a mouse interaction to inspect each operation. 

As practical comparison, we present the performance of our previous
and current state of the art FPGA implementation for latency
\cite{DAlbertoWNNDS2021}.

\subsection{Resnet 50}
\singlefigure{0.99}{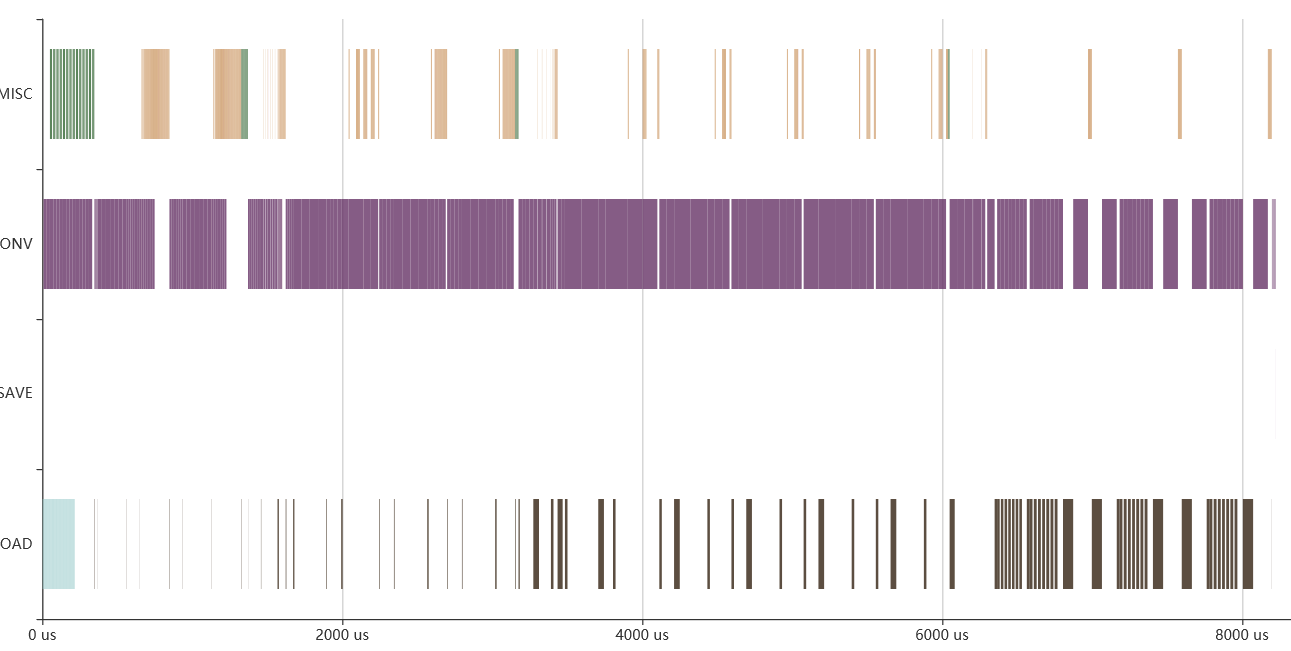}{Compiler Resnet Timing}{fig:resnet}

Figure \ref{fig:resnet} presents the visualization of Resnet50
assembly code generated by the compiler. This network is a reference
and every HW must do well. Briefly, the network expose the following
two basic optimizations:
\begin{equation}
  \label{eq:maxpool}
  \Vc{Y} = {\bf MaxPool} (\Vc{T} = \Tensor{\Vc{W}}{\Vc{X}} + \Vc{b})
\end{equation}
and
\begin{equation}
  \label{eq:eltwiseconv}
  \Vc{Y} =  (\Tensor{\Vc{W}}{\Vc{X}} + \Vc{b}) + \Vc{Z}
\end{equation}
with an explicit fusion operation will exploit parallelism between
convolution, pool, and element wise addition
\begin{equation}
  \label{eq:fus}
  \Vc{Y} =  \sum_{i=1}^{H/H_c}(ELTWISE(\Tensor{\Vc{W}}{\Vc{X}} + \Vc{b})_{H_c}, \Vc{Z}_{H_c})
\end{equation}
In Figure \ref{fig:resnet}, The first computation is a fused pipelined
convolution and maxpool where the input is in DDR. We can see how the
loads (light blue), convolutions (purple), and pool (green)
instructions are scheduled. Then there are two element wise addition
and convolution fusion and pipelining. At about the 1000 micro second
mark we can see the lag because of tail computation. The compiler at
this time will not start the next convolution because the output
tensor is not completely computed.
At the 6000 microsecond mark, we notice that
\begin{equation}
  \label{eq:tiling2}
  (\Tensor{\Vc{W}_T}{\Vc{X}})_{H}+\Vc{b}_T
\end{equation}
the compiler deploys the weight tiling and software pipelining: the
weights do not fit the PM memory and weight load instructions (black)
are pipelined with the convolution instructions.

The best performance we could achieve for resnet previously is 3.2ms
for one image \cite{DAlbertoWNNDS2021}.  The estimated and practical
execution time is 8.2 msec. The computation reach till the AVGPOOL
computation and it has a throughput of 4 images. The hand optimized
code is 7.5ms, Figure \ref{fig:manualresnet}.

\singlefigure{0.99}{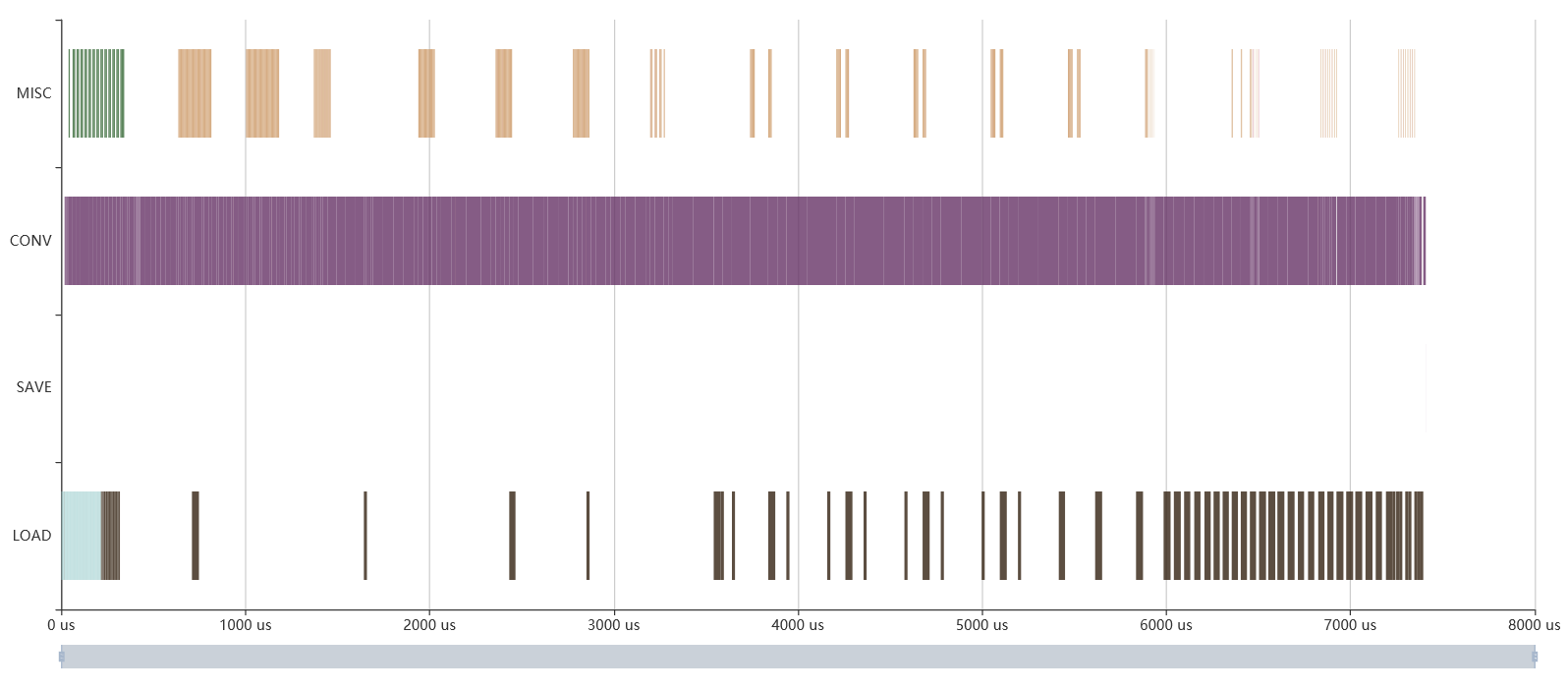}{Expert Resnet
  Timing}{fig:manualresnet}

As note, the hand tuned computation allows to starts the computation
of layers before the predecessor completion, because accurate timing
and clever use of circular buffers. Also a few max pools are removed
from the computation altogether by recognizing optimizations available
by the following 1x1 convolutions. The compiler does not deploy such a
specific optimizations because they are applicable only on this resnet50.

\subsection{Inception v1}
\singlefigure{0.99}{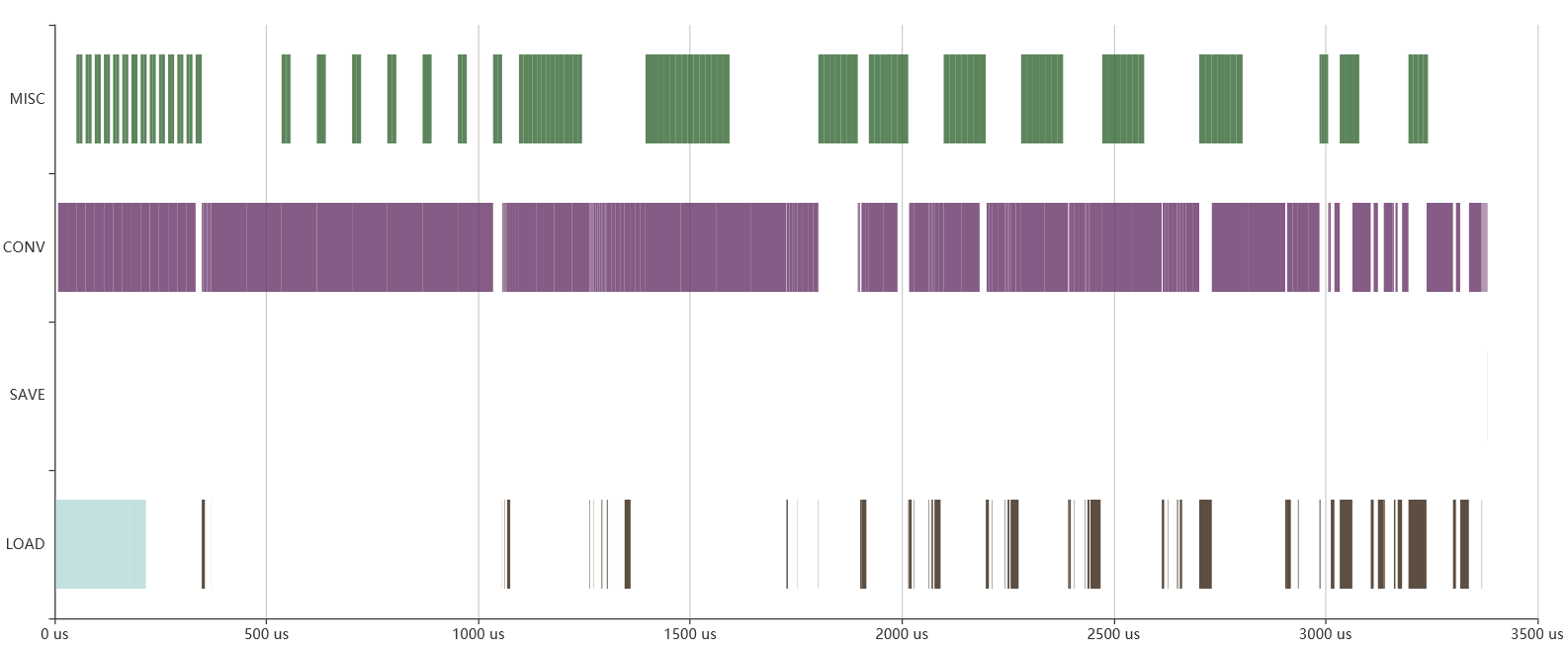}{Inception v1 Timing}{fig:inceptionv1}

Inception V1 is an example where concatenation-concatenation sections
of the network present two types of instruction parallelism. We can
execute in parallel a maxpool and a convolution, or a fusion of
convolution and maxpool (Equation \ref{eq:convpool}).  Although,
parallel node level instructions are preferable when possible (i.e.,
no head and not tail warm up and cool off), we exploit a simple
heuristic.  If the following convolution is the largest among the one
we could compute in parallel and the fusion allows a faster
implementation, we fuse the operations. Otherwise, we keep them
separated and let the scheduler seek for parallel schedule if
possible. At mark 1500 microsecond, we can see a successful parallel
scheduling. If you wonder about the pool (green) instructions at about
2000 microsecond, the max pool is a bottle neck of the computation and
we do not try to move it.

We can achieve 4 images every 3.4 ms, in our previous implementation
we achieve 1 image every 1 ms.

\subsection{Yolo v3}
\singlefigure{0.99}{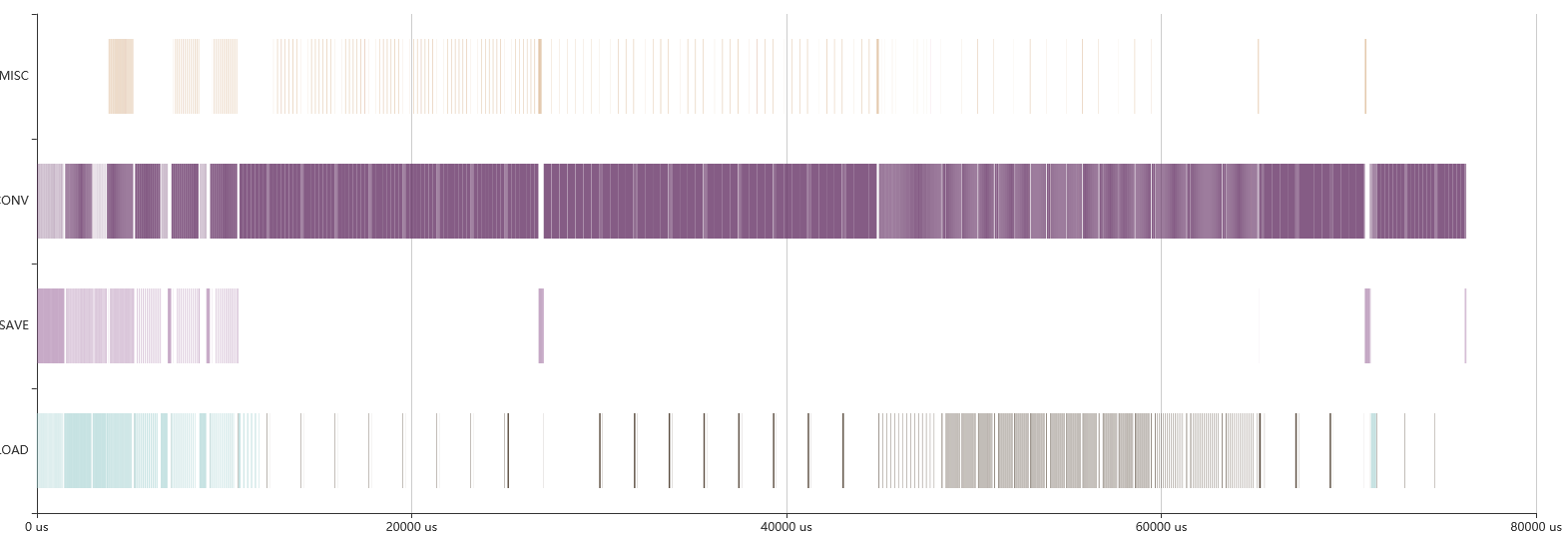}{Yolo V3 Timing}{fig:yolov3}

Yolo V3 has element-wise operation and concatenation instructions to
join different computation paths together (the combination of
inception and resnet). As previous networks, the last part of the
computation is weight heavy and the beginning is input/output
heavy. Differently there is more stress about memory allocation and
data movements. We need to use DDR as a swap space. Also a main
difference is that the size of the outputs is larger than the inputs. 

We can compute 4 images every 77 ms.

\subsection{VGG 16 (tensorflow)}
\singlefigure{0.99}{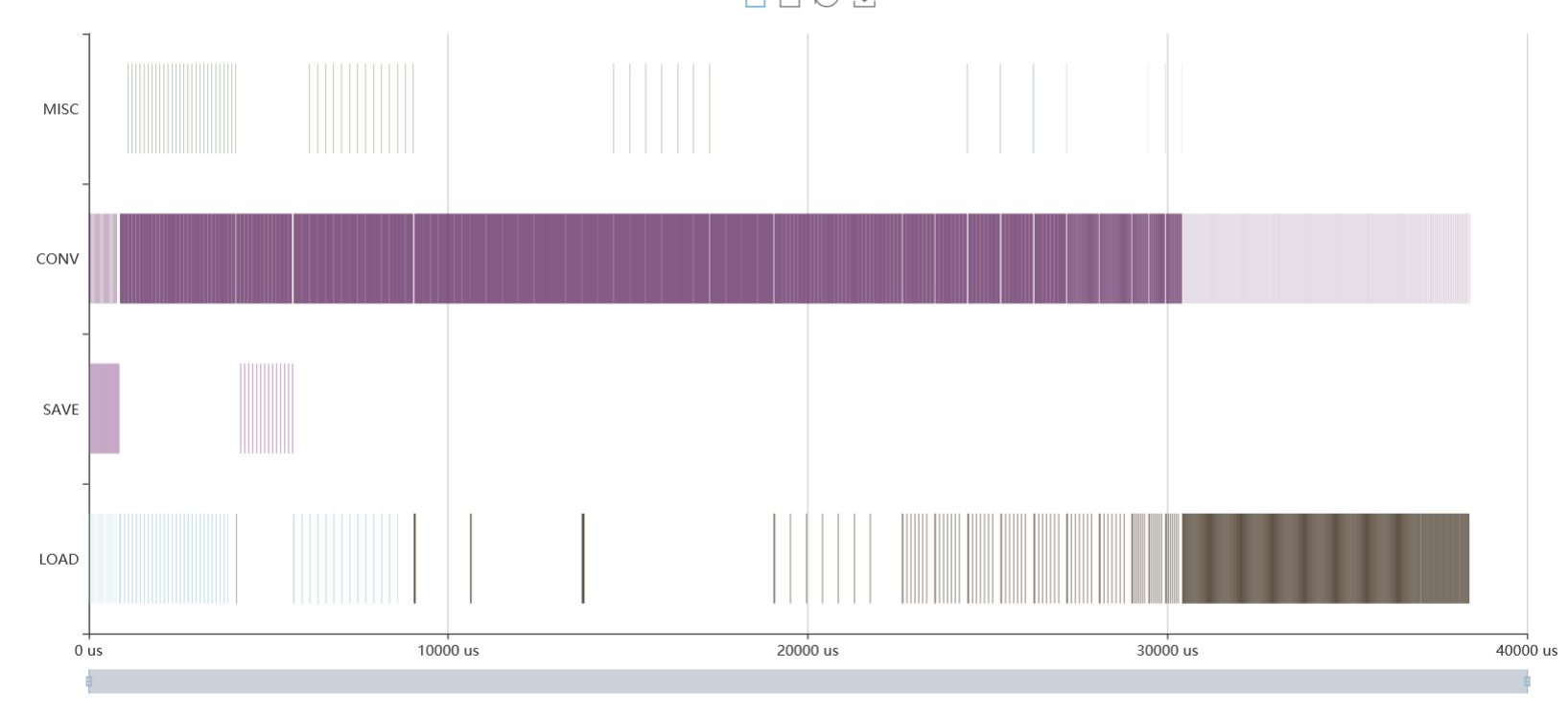}{VGG 16}{fig:vgg16}

VGG is a long sequence of heavy convolutions, and this version ends
the computation with three fully connected layers (FCs are ineffcient
for this architecture).

We can compute 4 images every 37.3 ms for VGG-16 and 44.5ms for
VGG-19, previously we could achieve 1 image every 18.2 ms and 20.5ms
respectively (for the caffe version we could do better VGG16 1 image
every 8.7ms using HW counters and 54ms considering the full runtime
stack wall clock, \cite{DAlbertoWNNDS2021}).

\subsection{Deconvolution and Cityscape}
\label{sec:deconvolutionp}

We start with a comparison of a synthetic deconvolution from a
citiscape model.
\singlefigure{0.99}{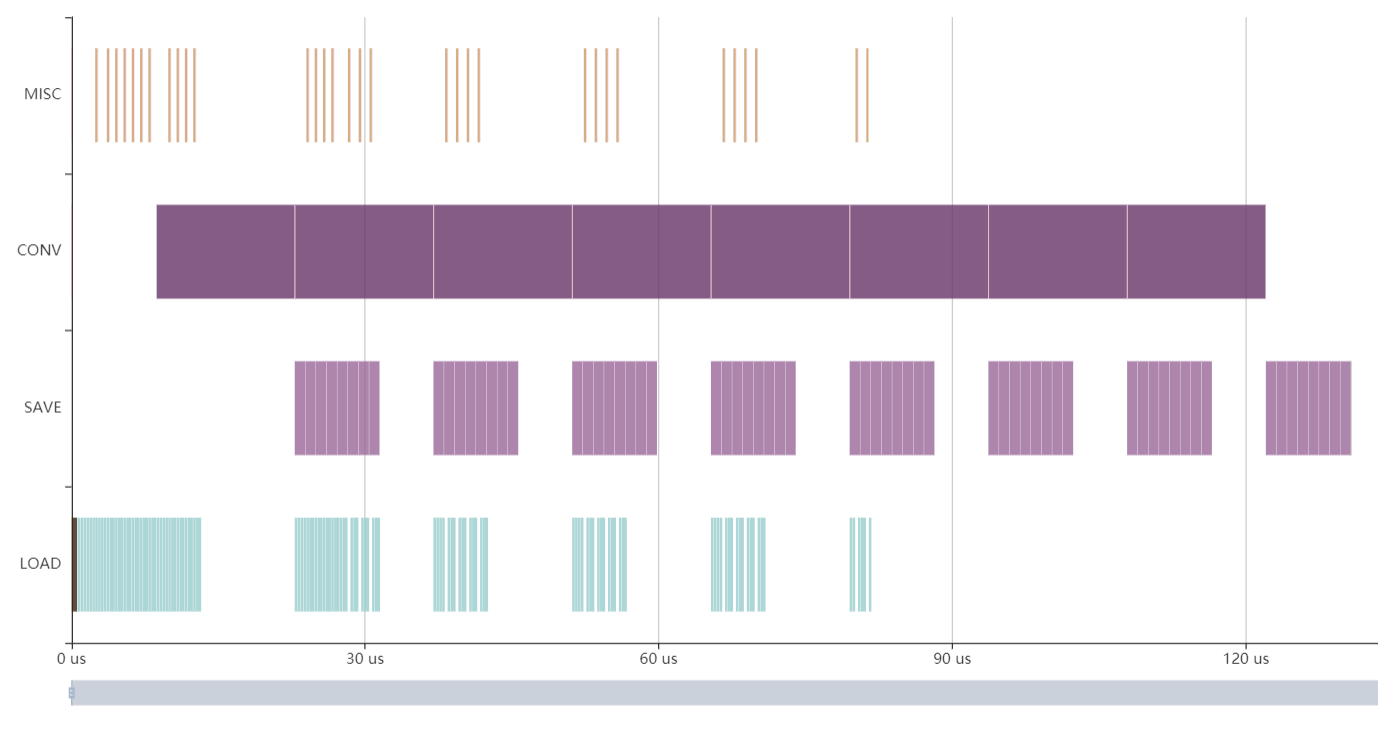}{Deconvolution =
  Load (blue) + Upsample (yellow + blue) + Convolution (dark purple) + Save (light purple)}{fig:deconv1}
In Figure \ref{fig:deconv1}, we present the classic
way to compute the deconvolution as a upsample and a Convolution. The
estimated execution time by simulation is 0.130ms and the practical
time is 0.55ms. The computation is convolution bound, this means that
the convolution dominates the overall execution time. The upsample is
a combination of data movement MISC operations and load from DDR,
where we store zeros in advance. In DPUV3INT8 2.0 the upsample is implemented using only MISC instructions.

In Figure \ref{fig:deconv2}, we have the optimal implementation using
only convolutions: the simulation time is 0.086ms and the practical
time is 0.42ms. The computation is dominated by storing the result in
DDR. These tests are single layer tests and thus reading and writing
in DDR are mandatory. Nonetheless, the execution-time advantages are
clearly measurable.

\singlefigure{0.99}{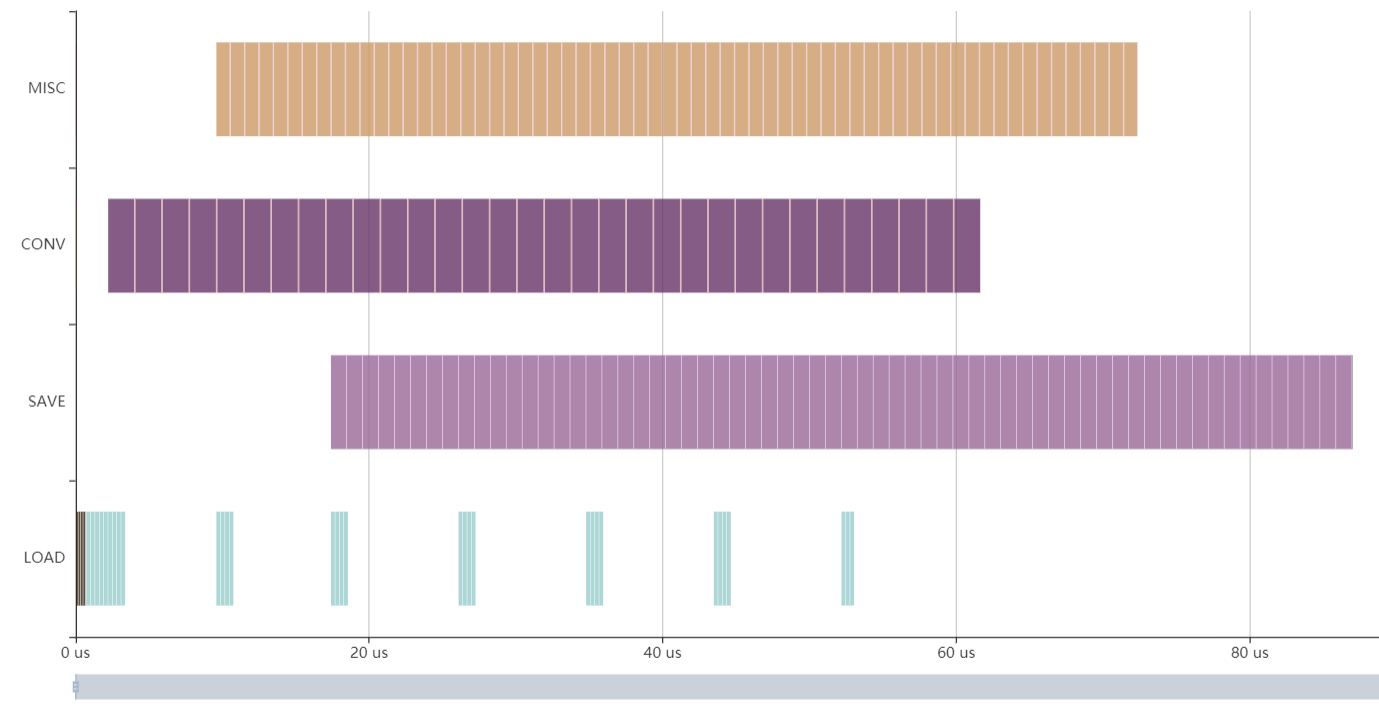}{Deconvolution = Load (blue) + Convolutions (dark purple) +
  Shuffle (yellow ) + Save (light purple)}{fig:deconv2}

\subsection{Model Zoo}
A model zoo is a collection of networks that is representative: it
provides variety of applications that can be applied to a variety of
hardware. This is a standardization. In particular, in the environment
we work every model in the zoo is compiled--executed using a common
interface. We call this interface {\em xcompiler}. The following
models are available in Vitis and they can run and tested
independently.

The xcompiler will take a model in native format, will create an
intermediate representation XIR and applies a set of standard
optimization passes. In short, a caffe model is transformed into a
xmodel (XIR) that we call xnnc xmodel. The xnnc is the front end
tool. The xcompiler takes the xnnc model and do optimization passes
(i.e., xcompiler xmodel), which is still an xmodel. The xcompiler
partitioner will take this xmodel, associate runners that execute CPU
subgraphs and FPGA subgraphs. In the previous sections, we presented
detailed performances for the xnnc models. Here we present
performances for the xcompiler partitioned xmodel, which is the
performance any user will tests as common interface in Vitis
\cite{vitis-ai}.

In this section, we present the execution time for the FPGA subgraph
(usually a single subgraph). We also present the estimated execution
time for the xnnc model (this has been used all along in the previous
sections and figures) and we provide also the estimated time for the
xcompiler xmodel. Being both XIR models the compiler is
able to generate code for either and thus we can estimate the
performance.

In Table \ref{tab:modelzoo}, we present the summary. Notice we can
provide a complete estimate set for xnnc models, a sub set for both
xcompiler xmodel and execution times. The missing measures are due to
different reasons in the creation of the different models.

\begin{table}[htb]
\caption{Model Zoo}
  \label{tab:modelzoo}
  \centering
  \begin{tabular}{|l|r|r|r|r|}
    \hline	
    Model          & Estimated       & Estimated           & Measured  \\
                   & xnnc            & partitioned         & latency   partitioned \\ 
    4 images       & (ms)            & (ms)                & (ms)      \\ \hline \hline
    vpgnet-pruned  & 5.447           & 5.447  &           \\
    plate-detect   & 1.021           & 1.243  &  3.168\\
    reid           & 1.527           & 1.747  &  2.497\\
    resnet-v1-50   & 8.158           & 9.005  &  9.519 \\ 
    resnet18       & 4.194           & 4.959  &  5.729\\
    resnet-v1-101  & 16.045          & 16.892 &  17.156\\ 
    resnet-v1-152  & 23.816          & 24.663 &  24.837\\ 
    inception-v3   & 37.588          & 37.788 &  38.379\\ 
    inception-v4   & 96.449          & 94.970 &  94.091 \\
    MLPerf-resnet50-v1.5  & 10.582   & 11.278 &  12.094\\
    vgg-16         & 37.499          & 37.515 &  37.024\\
    vgg-19         & 45.077          & 45.093 &  45.116\\
    yolov3-voc-tf  & 62.949          & 74.394 &  78.045\\
    densebox-320-320 & 1.006         & 1.228  &  3.056\\
    densebox-640-360 & 2.213         & 2.719  &  5.409\\
    yolov3-adas-pruned-0-9 & 3.449   & 7.153  &  10.1864\\
    inception-v1  & 4.470            & 5.422  &  5.720\\
    inception-v1 parallel            & 3.463  & TBD    &  3.507 \footnote{Debug mode no xcompiler partitioned} \\
    face-landmark & 0.540            & TBD    &  1.728\\
    resnet50      & 9.605            & 10.382 &  10.901\\
    tiny-yolov3-vmss & 4.939         & 6.526  &  NA \\
    yolov3-voc    & 62.949           & 74.394 &  NA \\
    openpose-pruned-0-3 & 185.334    & 185.617 &  184.071\\
    squeezenet    & 1.174            & 1.150  & NA  \\
    plate-num     & 3.048            & 5.777  &  4.356\\
    sp-net        & 1.093            & NA & NA  \\
    inception-v1-tf  & 4.402         & NA & NA \\
    inception-v1-tf parallel  & 3.463& NA & NA \\
    \hline
  \end{tabular}
\end{table}

In this section, we want to emphasize that xcompiler xmodels may have
more layers executed in FPGA than the xnnc xmodel. There are multiple
reasons:

\begin{itemize}
\item the optimization passes will transform more Fully Connected
  layer into convolutions
\item the xcompiler introduces down-sample operations
\item it may change slightly the network
  
\end{itemize}
Note: inception-v1 parallel we have 3.507ms execution time with NO
xcompiler partition.

\section{Conclusions}
FPGA solutions present an interesting scenario where using the same
device and different designs we can achieve very different latency
time and throughput.  In our work, we show how compilers have to rise
to the challenge: here we present the core optimizations to exploit
streaming and instruction parallelism and in so far this is the best
solution available in the Xilinx's DPU portfolio.

\bibliographystyle{IEEEtran}
\bibliography{ref}

\end{document}